\documentclass[letterpaper, 10 pt, conference]{ieeeconf}  

\IEEEoverridecommandlockouts                              
\usepackage{graphics} 
\usepackage{amsmath} 
\usepackage{amssymb}  
\usepackage[export]{adjustbox} 
\usepackage{hyperref}
\usepackage{cleveref}
\usepackage{mathtools}
\usepackage{bm}
\usepackage{subcaption} 
\usepackage{tikz}
\usepackage{booktabs}
\usepackage{xcolor}
\usepackage{graphics}
\usepackage{multirow}
\usepackage{float}

\usepackage[font=small]{caption}

\title{\LARGE \bf
RGB-D-Inertial SLAM in Indoor Dynamic Environments with Long-term Large Occlusion
}

\author{Ran Long$^{1,*}$, Christian Rauch$^{2}$, Tianwei Zhang$^{3}$, Vladimir Ivan$^{4}$, Tin Lun Lam$^{3,5}$, Sethu Vijayakumar$^{1,*}$%
\thanks{
This research is supported by the European Union’s Horizon 2020 research and innovation programme under grant agreement No 101017008 (Harmony) and the Alan Turing Institute.}%
\thanks{$^{1}$The authors are with the Institute of Perception, Action and Behaviour, School of Informatics, University of Edinburgh, Edinburgh, EH8 9AB, U.K.}%
\thanks{$^{2}$The author is with Bosch Center for Artificial Intelligence, Germany.}%
\thanks{$^{3}$The authors are with the Shenzhen Institute of Artificial Intelligence and Robotics for Society (AIRS), Shenzhen, China.}%
\thanks{$^{4}$The author is with the Touchlab Limited, U.K.}%
\thanks{$^{5}$The author is with the School of Science and Engineering, the Chinese University of Hong Kong, Shenzhen, China.}
\thanks{$^{*}$Corresponding authors: Ran Long (\texttt{Ran.Long@ed.ac.uk}), Sethu Vijayakumar (\texttt{Sethu.Vijayakumar@ed.ac.uk}).}%
}

\begin{document}

\maketitle
\thispagestyle{empty}
\pagestyle{empty}

\begin{abstract}
This work presents a novel RGB-D-inertial dynamic SLAM method that can enable accurate localisation when the majority of the camera view is occluded by multiple dynamic objects over a long period of time. Most dynamic SLAM approaches either remove dynamic objects as outliers when they account for a minor proportion of the visual input, or detect dynamic objects using semantic segmentation before camera tracking. Therefore, dynamic objects that cause large occlusions are difficult to detect without prior information. The remaining visual information from the static background is also not enough to support localisation when large occlusion lasts for a long period. To overcome these problems, our framework presents a robust visual-inertial bundle adjustment that simultaneously tracks camera, estimates cluster-wise dense segmentation of dynamic objects and maintains a static sparse map by combining dense and sparse features. The experiment results demonstrate that our method achieves promising localisation and object segmentation performance compared to other state-of-the-art methods in the scenario of long-term large occlusion.
\end{abstract}

\section{Introduction}
\begin{figure}[htb]
    \centering
    \setlength{\belowcaptionskip}{-0.5cm}
    \includegraphics[width=0.95\linewidth]{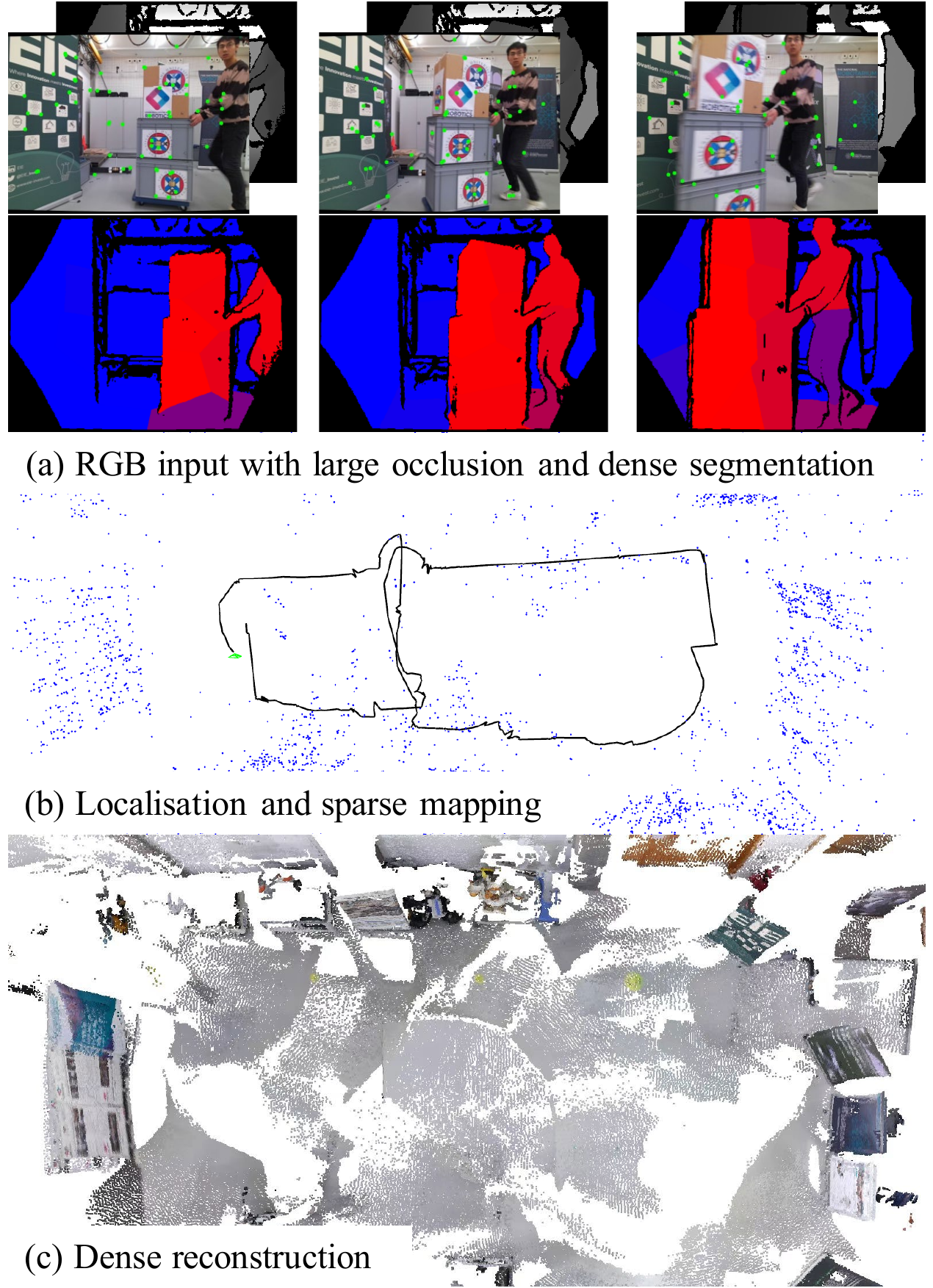}
    \caption{(a) In the scenario of long-term large occlusion, the majority of camera view is occluded for the majority of time frames. Our method can estimate cluster-wise dense segmentation of dynamic objects, and (b) simultaneously localise the camera and create a static sparse map. (c) The dense reconstruction of the static background can be acquired using the estimated camera trajectory and dense object segmentation after the procession of the whole sequence.}
    \label{fig:teaser}
\end{figure}

Simultaneous localisation and mapping (SLAM) is one of the core problems in various robot applications. Despite providing accurate camera motion estimation and mapping in large-scale environments, most existing SLAM systems \cite{mur2017orb, whelan2015elasticfusion} assume that the environment is static. This assumption can be violated when a robot closely manipulates objects in the scene or collaborates with other humans over a long period. In this scenario, the dynamic objects can cause long-term large occlusion, which means for the majority of time when a robot moves in the environment, the major proportion of the camera view is occluded by multiple dynamic objects.

To enable robust SLAM in dynamic environments, many visual SLAM methods \cite{scona2018staticfusion, runz2017co} detect the areas of dynamic objects by assuming that the static background occupies a major fraction of the camera view. The dynamic objects can, therefore, be removed as outliers during robust camera tracking. On the other hand, when the categories of dynamic objects are predefined, the regions containing these objects can be directly detected using deep learning methods \cite{ren2022visual}. 

However, when a priori undefined dynamic objects cause long-term large occlusion, there are still two major challenges. First, robots are unable to differentiate the dynamic objects from the static background because they can neither be detected by semantic segmentation nor be removed as outliers. Moreover, even when the dynamic objects are correctly removed, the remaining colour and depth information from the static background may be inadequate to support accurate localisation or mapping.

Various dynamic SLAM methods \cite{long2021rigidfusion, long2022pnpfusion, song2022dynavins} have explicitly considered dynamic objects that are dominant in the camera views with the aid of robot proprioception, like wheel odometry or Inertial Measurement Units (IMU). Some static visual-inertial navigation system (VINS) methods \cite{campos2021orb, stumberg2022dmvio} have also demonstrated their robustness when the camera view is fully occluded for a short period. However, none of them has shown their performance if the large occlusion lasts for the most of time when the camera is in motion. 

This paper is aimed to enable robust dynamic SLAM in the presence of long-term large occlusion. We use motion priors from IMU in a tightly-coupled way to help detect dynamic objects that cause large occlusion by simultaneously estimating camera motion, object segmentation and bias terms of the IMU. However, when the major proportion of camera view is occluded for a long period, the remaining features from the static background are unable to provide an accurate bias estimation of IMU which is important to detect dynamic objects that cause large occlusion. To further improve the method's robustness to long-term large occlusion, we actively remove sparse map points that are generated from the regions of dynamic objects and maintain a sparse model of the static background by integrating both sparse and dense features. The dense reconstruction of the static background is acquired after the processing of the whole sequence (\Cref{fig:teaser}). 

In summary, this work contributes:
\begin{enumerate}
    \item a novel methodology that combines sparse and dense features for dynamic object detection.
    
    \item a new bundle adjustment (BA) pipeline that simultaneously provides dense segmentation of dynamic objects, tracks the camera and maps the environments. 
    \item an RGB-D-inertial SLAM method that is robust to long-term large occlusion caused by multiple undefined dynamic objects.
\end{enumerate}

\section{Related Work}
\subsection{Visual Dynamic SLAM}
To detect regions of dynamic objects, one feasible solution is to use deep learning methods when object categories are provided a priori. EM-Fusion \cite{Strecke2019EMFusion} uses pixel-wise semantic segmentation provided by Mask R-CNN \cite{he2017mask} to directly detect dynamic objects. The problems of multiple object tracking (MOT) and SLAM are then integrated into an expectation maximisation (EM) framework. DynaSLAM \cite{bescos2018dynaslam} proposes to inpaint the dense model of the static background after dynamic object removal and, therefore, improves the accuracy and robustness of camera tracking. DP-SLAM \cite{li2021dp} further combines semantic segmentation with geometric constraints to refine dynamic object segmentation. However, these methods are limited to handling dynamic objects that are predefined in the deep learning methods. 

To remove undefined dynamic objects, StaticFusion (SF) \cite{scona2018staticfusion} assumes that the static background is dominant in the camera view and estimates the motion of the major rigid body. The dynamic objects are detected by their high residuals under the camera motion and then removed as outliers. Co-Fusion (CF) \cite{runz2017co} can further reconstruct the regions of outliers and track them independently. Similarly, MultiMotionFusion \cite{rauch2022mmf} can reconstruct dynamic objects online once they are detected and can re-detect them after they reappear in the camera view with the object models. Multimotion visual odometry (MVO) proposes an online multimotion pipeline to simultaneously estimate the segmentation of multiple dynamic objects and estimate their rigid motions independently. However, all these methods assume that the static background is the largest rigid body of motion and are, therefore, unable to detect dynamic objects that cause large occlusion.

\subsection{Proprioception-aided SLAM}
The combination of visual sensors and robot proprioception, like IMU and wheel odometry, can improve the robustness of SLAM in real environments. 

VINS-Mono \cite{qin2018vins} combines the measurements from a monocular camera and an IMU in a tightly-coupled manner and proposes an accurate visual-inertial (VI) SLAM system in large-scale environments. DM-VIO \cite{stumberg2022dmvio} delays the marginalisation of previous frames for a certain period to retain the prior information and improves the robustness of system. ORB-SLAM3 \cite{campos2021orb} is a real-time tightly-coupled VI SLAM system that maintains multiple maps simultaneously and can reuse all previous information from a co-visible graph when the camera revisits a place. Despite assuming the environment is static, these methods have achieved accurate localisation when the camera view is fully covered for a short period \cite{schubert2018tum}. However, this occlusion is caused by featureless objects, such as a dark tube, instead of large or closely moving dynamic objects with rich features.

To handle predefined dynamic objects, Qiu \textit{et al}. \cite{qiu2019tracking} integrates semantic bounding boxes with VI SLAM systems and enables object 3-D motion tracking from 2-D regions of images. Similarly, Dynamic-VINS \cite{Liu2022dynamicvins} refines 2D bounding boxes generated from YOLOv3 \cite{redmon2018yolov3} and removes feature points of dynamic objects with a recourse-limited platform. Ren \textit{et al}. \cite{ren2022visual} proposes a dense RGB-D-inertial SLAM system that can track and relocalise multiple dynamic objects with the aid of instance segmentation from Mask R-CNN \cite{he2017mask}. In contrast, DynaVINS \cite{song2022dynavins} can remove undefined dynamic objects that are dominant in the visual input using the camera motion priors from a low-cost IMU. It also actively detects temporarily static objects to reject false loop closure constraints. Our previous work RigidFusion \cite{long2021rigidfusion} uses camera motion prior from wheel odometry to simultaneously track the camera and a single rigid object that causes large occlusion, which is then extended to multiple planar object tracking \cite{long2022pnpfusion}. However, none of these methods has demonstrated their performance if the large occlusion in the camera view lasts for the majority of time when the robot perceives the world.

\section{Methodology}
\subsection{Overview and Notation}
\begin{figure*}[htb]
    \centering
    \setlength{\belowcaptionskip}{-0.5cm}
    \includegraphics[width=\linewidth]{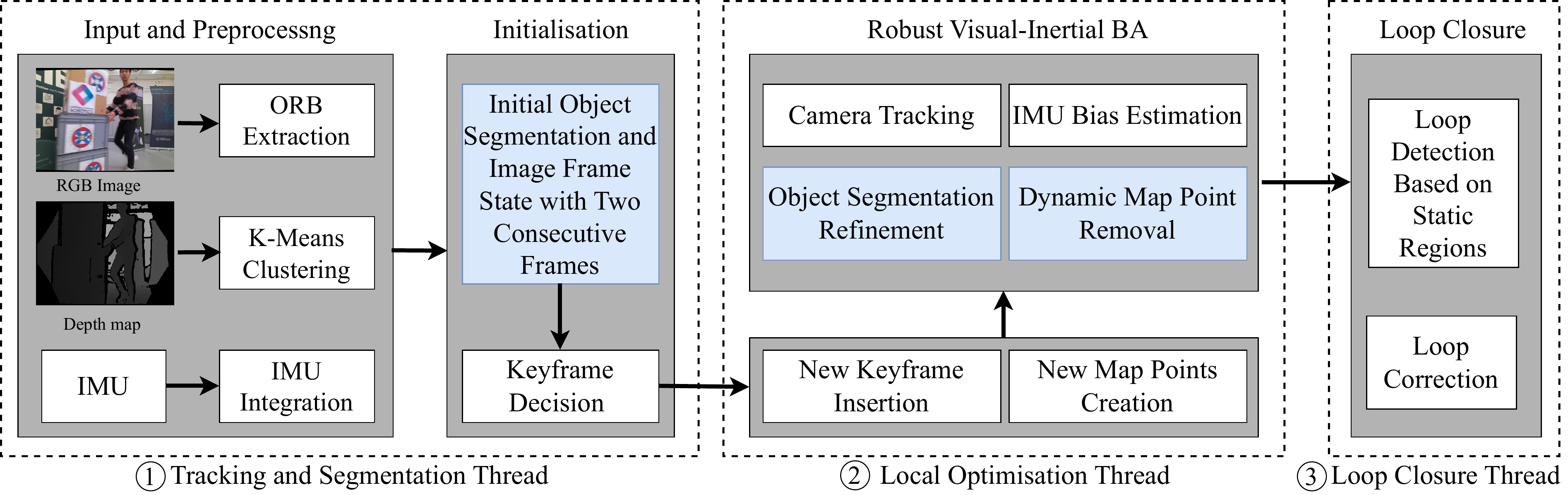}
    \caption{The pipeline of our method is based on ORB-SLAM3 \cite{campos2021orb} and blue rectangles highlight the functions we implement in addition to ORB-SLAM3. Our pipeline consists of three threads: (1) In the \textit{tracking and segmentation} thread, we extract ORB features \cite{rublee2011orb} from colour images and over-segment the images into clusters by applying K-Means clustering on the depth image. Given IMU bias estimation, we acquire camera motion priors using pre-integrated velocity, rotation and position measurements. We then estimate initial object cluster-wise segmentation and image frame states based on a combination of sparse and dense features. (2) In the \textit{local optimisation} thread, new keyframes are created and sparse map points are generated from the initial static parts of the image. We then conduct robust visual-inertial BA to simultaneously remove dynamic map points, estimate the states of multiple keyframes and refine dense object segmentation. (3) Finally, the static parts of keyframes are used for place recognition and loop closure in the \textit{loop closure} thread. }
    \label{fig:pipeline}
\end{figure*}

The proposed new RGB-D-inertial SLAM pipeline (\Cref{fig:pipeline}) focuses on accurate localisation and dense segmentation of dynamic objects that cause long-term large occlusion. Our method takes a stream of RGB-D images and readings from a low-cost IMU as input, which means additional variables are needed to be estimated compared to visual SLAM. For a frame $i$, similar to ORB-SLAM3 \cite{campos2021orb}, we consider the pose $T_i \in SE(3)$ and velocity $v_i$ of body in the world frame, and the bias terms of gyroscope $\mathbf{b}^g_i \in \mathbb{R}^3$ and accelerometer $\mathbf{b}^a_i \in \mathbb{R}^3$ respectively. The state vector for frame $i$ is denoted as $S_i = \{T_i, v_i, \mathbf{b}^g_i, \mathbf{b}^a_i\}$. The camera pose $T_{Ci}$ can be acquired by $T_{Ci} = T_i T_{BC}$, where $T_{BC}$ is calibrated a priori and denotes the rigid transformation between the body (IMU) and camera frame.

For the image frame $i$, we extract ORB features \cite{rublee2011orb} from the intensity image $I_i$ at the feature point pixel location $\mathbf{u} \coloneqq (u, v) \in \mathbb{R}^2$. Feature points with a valid depth reading are treated as stereo feature points and the stereo coordinate $\mathbf{u}^s \in \mathbb{R}^3$ is defined as $[\mathbf{u}, u-f_xb/d]^T$, where $f_x$ is the focal length, $d$ is the depth and $b$ is the baseline for RGB-D cameras \cite{mur2017orb}. We also over-segment the depth image into $K$ clusters and the region with no valid depth reading is denoted as the $(K+1)$th cluster. For each cluster, we assign a score $\gamma \in [0,1]$ to represent the probability that the cluster is static. The whole set of scores is denoted as $\Gamma_i = \{\gamma_{i0}, \cdots, \gamma_{iK-1}, \gamma_{iK}\}$, where $\gamma_{iK} = 1$. To reduce computation complexity, we assume all pixels and feature points in the same cluster have the same score. Concretely, $\gamma(\mathbf{u}^s) \coloneqq \gamma(\mathbf{u})$ denotes the score of the pixel $\mathbf{u}$ and is equal to $\gamma_{ik}$ for any pixel $\mathbf{u}$ that belongs to the $k$-th cluster at the $i$-th image frame.  

Between two consecutive images $i-1$ and $i$, we estimate pre-integrated measurements of rotation $\Delta \mathbf{R}_{i-1,i}$, velocity $\Delta \mathbf{v}_{i-1,i}$ and position $\Delta \mathbf{p}_{i-1,i}$ from their states $S_{i-1}$, $S_i$ based on the theory proposed in \cite{lupton2011visual, forster2016manifold} and denote the inertial residual $\mathbf{r}^{I}_{i-1,i}$ as $[\Delta \mathbf{R}_{i-1,i}, \Delta \mathbf{v}_{i-1,i}, \Delta \mathbf{p}_{i-1,i}]$.

Our robust BA considers a set of $M+1$ co-visible keyframes within a sliding window and a set of $L$ 3-D map points that are observed by these keyframes \cite{campos2021orb}. The states of keyframes are denoted as $\hat{S}=\{S_0, \cdots, S_M\}$ and the 3-D positions of map points are denoted as $\hat{X} = \{\mathbf{x}_0, \cdots, \mathbf{x}_{L-1}\}$. For all keyframes, the set of cluster-wise dense segmentation is denoted as $\hat{\Gamma} = \{\Gamma_0, \cdots, \Gamma_{M}\}$. In addition, for each map point $j$, we also assign a score $\beta_j \in [0,1]$ to represent the probability that the map point is generated from the regions of the static background. The scores of all map points are denoted as $\hat{B} = \{\beta_0, \cdots, \beta_{L-1}\}$.

\subsection{Robust Visual-inertial Bundle Adjustment (BA)}
To handle long-term large occlusion caused by dynamic objects, we simultaneously estimate the segmentation of dense images and sparse map points, track the camera motions and correct IMU biases. The static map points are used to improve the consistency and robustness of object dense segmentation over a long period. To achieve it, we propose a novel energy function that consists of four energy terms:
\begin{equation}
\begin{aligned}
\label{eq:viba}
    &\min_{\hat{S}, \hat{\Gamma}, \hat{B}} \left(
    U(\hat{S}) + R(\hat{S}, \hat{\Gamma}, \hat{B}) + G(\hat{\Gamma}, \hat{B}) + H(\hat{S}, \hat{\Gamma}) \right),\\
    &\text{ s.t.\ $  \gamma_{ik}, \beta_j \in [0,1] \  \forall i, j,k$},
\end{aligned}
\end{equation}
where the first term $U(\hat{S}) = \sum_{i=1}^{M} ||\mathbf{r}^I_{i-1,i}||^2_{\Sigma^{-1}_{i-1,i}}$ is the IMU residual term \cite{campos2021orb} which provides motion priors for any two consecutive keyframes. 

The second term $R(\hat{S}, \hat{\Gamma}, \hat{B})$ is the robust residual error between map points and keyframes and is defined as:
\begin{equation}
\label{eq:rr}
     \sum_{j=0}^{L-1} \sum_{i \in \mathcal{K}^j}\beta_j^2 \gamma(\mathbf{u}_{ij}^s)^2 \rho_H(||\mathbf{r}_{ij}||_{\Sigma_{ij}^{-1}}) + (1-\beta_j)^2 (1-\gamma(\mathbf{u}_{ij}^s))^2 \hat{c},
\end{equation}
where $\mathcal{K}^j$ includes all keyframes that observe map point $j$, $\mathbf{u}_{ij}$ represents the observation of map point $j$ at the keyframe $i$ and $\gamma(\mathbf{u}_{ij})$ gives the probability that this observed feature point belongs to the static background. The robust Huber loss $\rho_H(\cdot)$ is used to reduce the effect of outliers \cite{mur2017orb} and $\hat{c}$ is a threshold. $\mathbf{r}_{ij} \in \mathbb{R}^3$ denotes the residual between map point $j$ and observation $\mathbf{u}_{ij}$:
\begin{align}
    \mathbf{r}_{ij} = \mathbf{u}_{ij}^s - \pi_s (T_{Ci} \mathbf{x}_j),
\end{align}
where $\pi_s$ is the projection function described in \cite{mur2017orb} for stereo keypoints. The robust residual item in \Cref{eq:rr} is inspired by DynaVINS \cite{song2022dynavins}, but the main difference is that we separately consider the dynamic segmentation of image feature points and map points.

The third term $G(\hat{\Gamma}, \hat{B})$ provides prior information for the cluster-wise dense segmentation $\hat{\Gamma}$ and sparse segmentation $\hat{B}$, and shows how $\hat{\Gamma}$ and $\hat{B}$ affect each other:
\begin{align}
\label{eq:segPrior}
    G(\hat{\Gamma}, \hat{B}) = \lambda_{\gamma} \sum_{i=1}^{M}\sum_{k=0}^{K-1}(\gamma_{ik} - \tilde{\gamma}_{ik})^2 + \lambda_{\beta} \sum_{i=0}^{L-1} (\beta_i - \tilde{\beta}_i)^2,
\end{align}
where $\tilde{\gamma}$ and $\tilde{\beta}$ are the priors of $\gamma$ and $\beta$ respectively, $\lambda_{\gamma}$ and $\lambda_{\beta}$ are parameters to weight the two terms. Inspired by StaticFusion \cite{scona2018staticfusion}, we assume that the dynamic regions of the depth map have a large depth difference from the static map. However, in contrast to StaticFusion, we estimate the depth difference between sparse map points and dense depth images. The $\tilde{\gamma}$ is, therefore, defined as:

\begin{align}
    \tilde{\gamma}_{ik} = 1 - \lambda_p \frac{\sum_{j \in \mathcal{M}^{ik}} |\beta_j(D_i(\pi(T_{Ci}\mathbf{x}_j)) - |T_{Ci}\mathbf{x}_j|_z)|}{n(\mathcal{M}^{ik})},
\end{align}
where $\mathcal{M}^{ik}$ is a set of map points that are observed by the $k$th cluster of the $i$-th keyframe and $\pi: \mathbb{R}^3 \rightarrow \mathbb{R}^2$ is the projection function of a pinhole camera. $|\cdot|_z$ gives the z-coordinate of a 3-D vector and $n(\cdot)$ provides the number of elements in a set. $\lambda_p$ is a parameter to control the influence of the depth difference to $\tilde{\gamma}$. 

Additionally, we assume that a map point belongs to the static background if it is classified as static for most of keyframes that can observe this map point:
\begin{align}
    \tilde{\beta}_j = \phi \left(\lambda_{\phi}, \frac{\sum_{i\in \mathcal{K}^j} \gamma(\mathbf{u}_{ij})  }{n(\mathcal{K}^j)} \right),
\end{align}
where $\mathcal{K}^j$ is defined in \Cref{eq:rr}. $\phi(\lambda_{\phi}, x)$ denotes $max(0, \frac{x-\lambda_{\phi}}{1-\lambda_{\phi}})$, which is inspired by ReLU \cite{agarap2018deep}. $\lambda_{\phi} \in [0,1)$ is a parameter and is chosen as 0.5 in implementation, which means a map point $j$ is dynamic if more than 50\% of keyframes in $\mathcal{K}^j$ classifies the map point $j$ as dynamic. 

The last term $H(\hat{S}, \hat{\Gamma})$ maintains the smoothness of dense segmentation $\hat{\Gamma}$ and adds regularisation on the IMU biases:

\begin{align}
\label{eq:regu}
    \sum_{i=1}^{M} \left( \lambda_r E(\Gamma_i) + ||\mathbf{b}_i^g - \mathbf{b}_{i-1}^g||^2_{\Sigma_{bg}}+||\mathbf{b}_i^a - \mathbf{b}_{i-1}^a||^2_{\Sigma_{ba}} \right).
\end{align}
This means for an individual keyframe $i$, we encourage neighbouring clusters to have close scores:

\begin{align}
    E(\Gamma_i) = \sum_{(j, k) \in V_i}(\gamma_{ij} - \gamma_{ik})^2 \ ,
    \label{equ::sspace}
\end{align}
where $V_i$ is the connectivity graph for clusters \cite{scona2018staticfusion} in the $i$-th keyframe and $(j, k) \in V_i$ represents the $j$-th and $k$-th clusters of keyframe $i$ are connected in space. In addition, following \cite{forster2016manifold}, we assume that the IMU biases are changing slowly over time and can be modelled with a ``Brownian motion''. We, therefore, penalise the difference of IMU biases $\mathbf{b}^a$ and $\mathbf{b}^g$ for any two consecutive keyframes.

The novelty of formulation \ref{eq:viba} is that we actively estimate dense segmentation of input images $\hat{\Gamma}$ and sparse segmentation of map points using a combination of dense and sparse features. This is different to StaticFusion \cite{scona2018staticfusion} or PlanarFusion \cite{long2022pnpfusion} which only consider the segmentation of images. We can, therefore, recover a static sparse map and correct IMU biases even if the initial dense segmentation of images are unreliable. The static map and corrected IMU biases can then be used to aid dense segmentation of images in the presence of long-term large occlusion.

To solve the optimisation for a large map, we follow ORB-SLAM3 \cite{campos2021orb} and consider a sliding window of keyframes with their corresponding map points. We also incorporate observations of these points from covisible keyframes. To optimise \Cref{eq:viba}, we decouple the state of keyframes $\hat{S}$, and segmentation $\hat{\Gamma}$ and $\hat{B}$. Specifically, we first initialise the state and dense segmentation of the latest keyframe and set $\beta=1$ for all map points. For each iteration, we fix $\hat{\Gamma}$ and $\hat{B}$ while finding the optimal states $\hat{S}$ for $M+1$ keyframes. Then, we fix $\hat{S}$, while $\hat{\Gamma}$ and $\hat{B}$ are iteratively optimised by fixing one and analytically solving the other. After optimisation, we remove all dynamic map points.

\subsection{Initialisation of Segmentation and Image Frame State}
For every two consecutive frames $i-1$ and $i$, given the state of the previous frame $S_{i-1}$, we estimate the state of current frame $S_i$ and a cluster-wise dense segmentation $\Gamma_i$ of dynamic objects. If the frame is selected as a keyframe, we use the segmentation and state as the initialisation of the robust BA. The initialisation is a special case of our robust BA such that \Cref{eq:viba} is a multiple-frame-to-model optimisation while \Cref{eq:initial} is a single-frame-to-frame optimisation. Therefore, we similarly propose to minimise an energy function with four terms:
\begin{equation}
\begin{aligned}
\label{eq:initial}
    &\min_{S_i, \Gamma_i}  
    U_{ini}(S_i)+ R_{ini}(S_i, \Gamma_i) + G_{ini}(\Gamma_i) + H_{ini} (S_i, \Gamma_i), \\
    &\text{ s.t.\ $  \gamma_{ik} \in [0,1] \  \forall k$},
\end{aligned}
\end{equation}
where the first term $U_{ini}(S_i)$ is similarly the IMU residual term. However, the second term $R_{ini}(S_i, \Gamma_i)$ minimises the residuals between the two consecutive dense intensity and depth image pairs ($I_{t-1}, D_{t-1}$) and ($I_t, D_t$) so that features from texture-less areas can be considered -- a crucial departure from \Cref{eq:rr}. Following \cite{scona2018staticfusion, long2021rigidfusion}, $R_{ini}(S_i, \Gamma_i)$ is defined as a weighted sum of intensity and depth residuals:
\begin{align}
\label{eq:denseResidual}
    \sum_{\mathbf{u} \in U_i} \gamma(\mathbf{u})[F(\alpha_I w_{I}^{\mathbf{u}}r_{I}^{\mathbf{u}}(\Delta T_{Ci})) + F(w_{D}^{\mathbf{u}}r_{D}^{\mathbf{u}}(\Delta T_{Ci}))] \ ,
\end{align}
where $U_i$ is the set of pixels with a valid depth reading at the current image frame $i$ and $\mathbf{u} \in \mathbb{R}^2$ is the pixel coordinate. $\Delta T_{Ci} = (T_{i-1}T_{BC})^{-1}T_{i}T_{BC}$ denotes the relative camera pose between two consecutive frames. Given a pixel coordinate $\mathbf{u}$, the intensity residual $r_{I}^{\mathbf{u}}(T)$ and depth residual $r_{D}^{\mathbf{u}}(T)$ under a transformation $T \in SE(3)$ can be acquired by:
\begin{align}
    r_I^\mathbf{u}(T) &= I_{i-1}\left(\mathcal{W}(\mathbf{u}, T)\right) - I_i\left(\mathbf{u}\right)\\
    r_D^\mathbf{u}(T) &= D_{i-1}\left(\mathcal{W}(\mathbf{u}, T)\right) - |T \pi^{-1}(\mathbf{u}, D_i\left(\mathbf{u})\right)|_z \ ,
\end{align}
where $\mathcal{W}$ is the pixel-wise image warping function:

\begin{align}
    \mathcal{W}(\mathbf{u}, T) = \pi\left(T\pi^{-1}(\mathbf{u}, D_t(\mathbf{u}))\right).
\end{align}
In \Cref{eq:denseResidual}, parameters $w_{I}$ and $w_{D}$ are used to weight the residuals of intensity and depth respectively. The parameter $\alpha_I$ controls the scale of intensity residuals. We use Cauchy robust penalty:
\begin{equation}
    F(r) = \frac{c^2}{2} \log\left(1 + \left(\frac{r}{c}^2\right)\right)
\end{equation}
to improve the robustness of residual minimisation and $c$ is the inflection point of $F(r)$.

The third term of \Cref{eq:denseResidual} is a simplified version of \Cref{eq:segPrior} and $G_{ini}(\Gamma_i) = \lambda_\gamma \sum_{k=0}^{K-1} (\gamma_{ik} - \tilde{\gamma}_{ik})^2$. This is because we only consider the dense segmentation $\Gamma_i$ of the $i$-th frame and treat all map points as static: $\beta_j = 1, \forall j$. Similarly, the last term $H_{ini}(S_i, \Gamma_i)$ can be acquired by assigning $M=1$ in \Cref{eq:regu}. Compared to \cite{scona2018staticfusion, long2021rigidfusion}, we use pre-integrated IMU measurements and combine sparse and dense features to help dynamic object segmentation and camera tracking.

A coarse-to-fine scheme similar to StaticFusion \cite{scona2018staticfusion} is applied to align dense intensity or depth images in the solver of \Cref{eq:initial}. Concretely, for each incoming RGB-D image pair, we create image pyramids for both intensity and dense images by iteratively resizing the image to the half-size of the previous level. We start the minimisation from the coarsest level and initialise the next level using the intermediate results from the current level. In addition, the frame state $S_i$ and dense segmentation $\Gamma$ are decoupled in the solver. For each iteration, we first fix the dense segmentation $\Gamma_i$ and optimise $S_i$. We then find an analytical solution of $\Gamma_i$ given the value of $S_i$.

\subsection{Place Recognition and Loop Closing}
We adopt a similar place recognition policy to ORB-SLAM3 \cite{campos2021orb} and use the DBoW2 place recogniser \cite{mur2014fast} to detect loop candidates based on their appearance. Because of large occlusion, to improve the accuracy of place recognition, we remove the dynamic regions of each keyframe and only consider candidate keyframes when the region of static background is more than 80\% of the whole image.

After verification of loop closure matches, we conduct a full vision-only BA based on the static parts of keyframes and all static map points to reduce long-term drift.

\section{Evaluation}
\subsection{Setup}
\begin{figure}[htb]
    \centering
    \setlength{\belowcaptionskip}{0cm}
    \newcommand{\h}{3.4cm}
    \subfloat[Azure Kinect DK]{\includegraphics[height=\h]{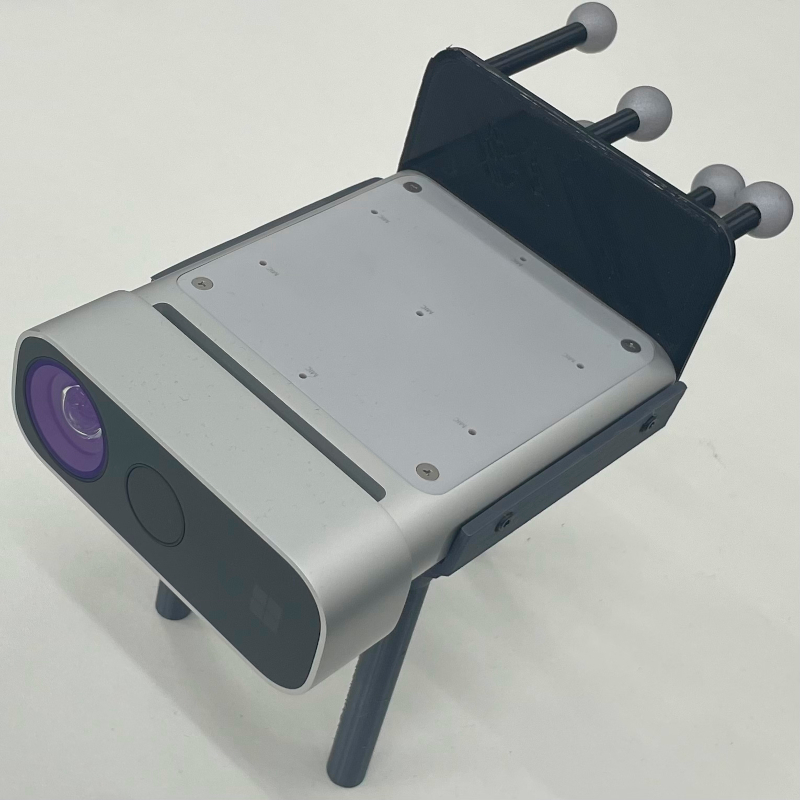} \label{fig:azure}}
    \subfloat[Wheeled Robot]{\includegraphics[height=\h]{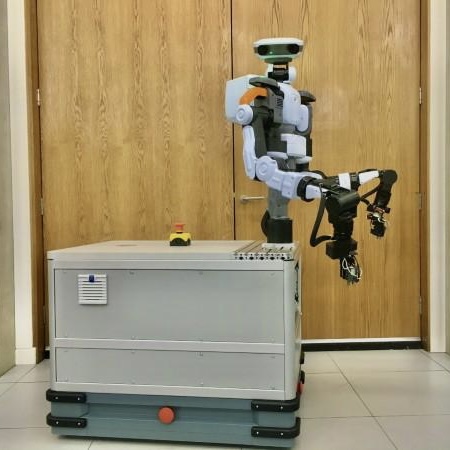} \label{fig:chonk}}
    \subfloat[Boxes]{\includegraphics[height=\h]{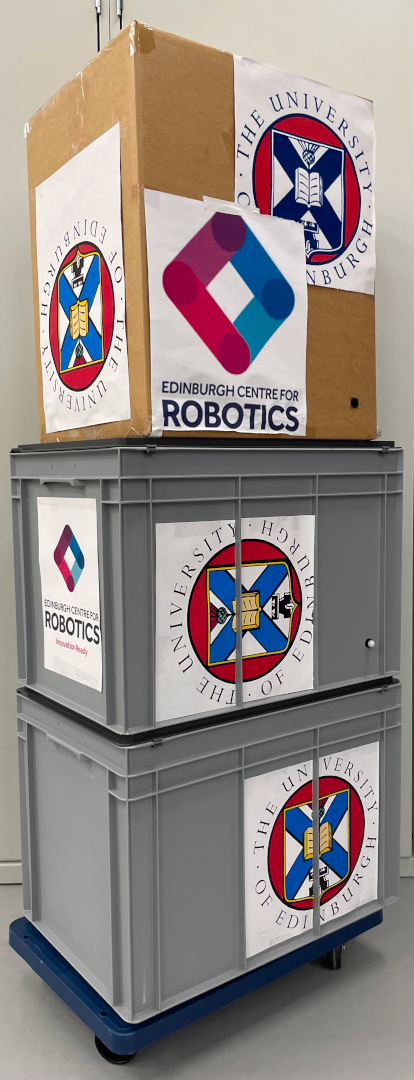} \label{fig:box}}
    \caption{(a) An Azure Kinect DK RGB-D camera with attached Vicon markers. (b) The base of an omnidirectional wheeled mobile manipulator on which the camera is mounted. (c) A large rigid object that can be moved by humans to cause large occlusion in the camera view.}
    \label{fig:setup}
\end{figure}

We collect evaluation sequences using an Azure Kinect DK RGB-D camera with an embedded low-cost IMU
(\Cref{fig:azure})
. The ground truth trajectories are acquired through Vicon system by attaching Vicon markers to the camera. The Azure Kinect DK can generate registered 1280 $\times$ 720 RGB-D image pairs at the frequency of 30 Hz and IMU readings at around 1700 Hz. To speed up the processing of image frames, we down-scale and crop the RGB-D images to the resolution of 640 x 480 (VGA). In the solver of \Cref{eq:initial}, the images are further down-scaled to 320 x 240 (QVGA) because of dense image alignment. 

During data collection, the camera is mounted on an omnidirectional wheeled robot 
(\Cref{fig:chonk}) 
and a human moves the stacked boxes 
(\Cref{fig:box}) 
closely in front of the camera to create large occlusion in the camera view. We collect nine sequences with different camera and object trajectories and they can be categorised into three types (\Cref{tab:sequence}) based on the proportion of large occlusion (LO) duration to the whole sequence: short-term (ST), mid-term (MT) and long-term (LT). For each category, we visualise the camera trajectory of one typical collected sequence and highlight the trajectory in red when the camera view is occluded by dynamic objects (\Cref{fig:motion}). For example, in \textit{seq7}, the majority of camera view is occluded for more than 70\% of image frames and the camera travels around 20 meters during this period, which increases the difficulty to localise camera. For quantitative evaluation, we follow \cite{sturm2012benchmark} and calculate the absolute trajectory error (ATE) and the relative pose error (RPE) against the ground truth camera trajectories.

\begin{table}[tb]
\setlength{\belowcaptionskip}{-0.8cm}
  \centering
  \resizebox{\columnwidth}{!}{%
    \begin{tabular}{|c|c|c|c|c|c|}
    \hline
          & Types & Dis. (m) & LO Dis. (m) & Durat. (s) & LO Durat. (s)\\
\hline
1 & Static  &  22.1   &   0 (0\%)   &    108   &  0 (0\%) \\
    \hline
    2     & \multirow{2}[1]{*}{ST} &  15.6    &   3.09 (19.8\%)    &    58.1   &  10.2 (17.6\%) \\
\cline{1-1}\cline{3-6}   
3   &    & 19.3   &   3.24 (16.8\%)    &    59.0   &      9.45 (16.0\%)  \\
    \hline
    4     & \multirow{2}[1]{*}{MT} &   17.2    &    5.83 (33.9\%)   &    49.9   &  14.7 (29.5\%)\\
\cline{1-1}\cline{3-6}    5     &   & 17.5   &   6.01 (34.3\%)    &   62.1    &  18.6 (30.0\%)  \\
    \hline
    6     & \multirow{4}[1]{*}{LT} &   21.9    &   14.9 (68.0\%)    &    85.8   &  48.7 (56.8\%)\\
\cline{1-1}\cline{3-6}    7     &       &   26.5    &   20.1 (75.9\%)    &    131   &  97.5 (74.4\%)\\
\cline{1-1}\cline{3-6}    8     &       &   34.1    &    21.5 (63.1\%)   &    189   & 101 (53.4\%) \\
\cline{1-1}\cline{3-6}    9     &       &    27.1   &    21.2 (78.2\%)   &    140   &  102 (72.9\%)\\
    \hline
    \end{tabular}%
    }
\caption{Statistics of nine collected sequences. ``Static'' means there is no dynamic objects in this sequence. Large occlusion (LO) distance or duration represents the distance or duration when the camera view is occluded respectively. Specifically, LTLO means the duration of large occlusion is longer than 50\% of the whole sequence duration.}
\label{tab:sequence}%
\end{table}%

\begin{figure*}[htb]
    \setlength{\belowcaptionskip}{0cm}
    \centering
    \includegraphics[width=\linewidth]{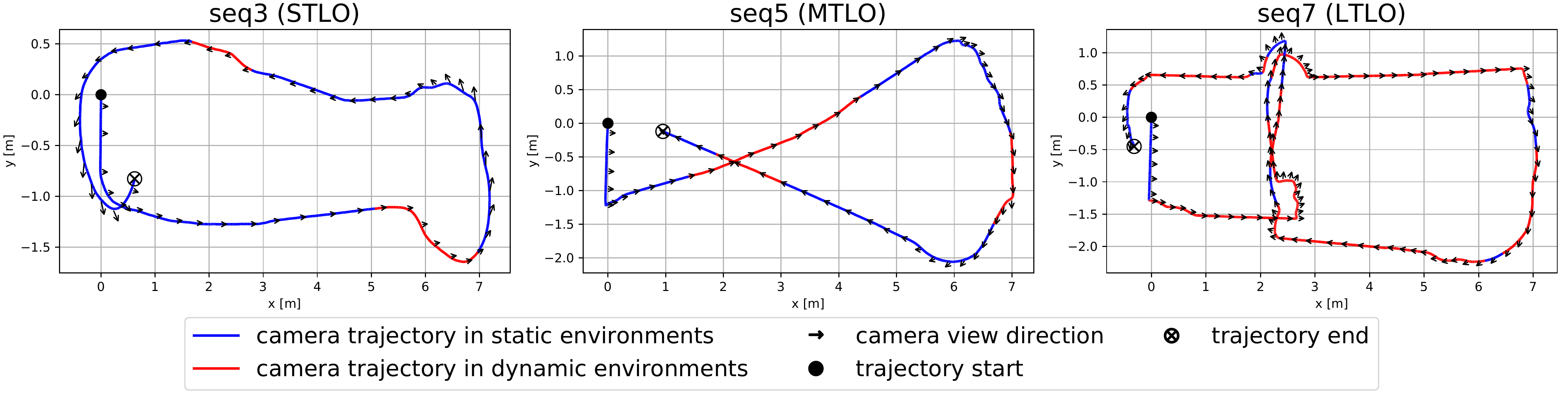}
    \caption{The camera ground truth trajectories from top-down perspective. The blue trajectory segment illustrates the part when there are no moving objects in the camera view. While the red segment represents that dynamic objects can be observed in the camera view. The start position of a trajectory is marked with a black solid dot and the end position is marked with a circle-cross marker. Finally, the black arrows point in the direction of camera view.}
    \label{fig:motion}
\end{figure*}

\subsection{Camera Localisation}
\begin{figure*}[htb]
    \centering
    \includegraphics[width=\linewidth]{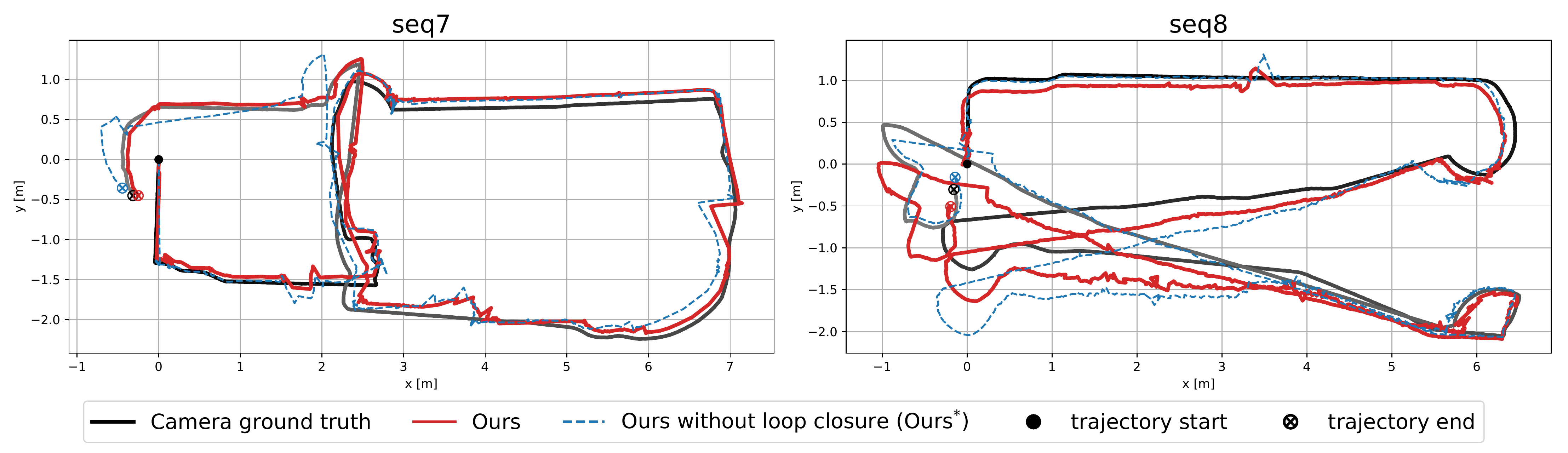}
    \caption{Visualisation of the estimated camera trajectories compared with the ground truth. We align the start position of all trajectory to the same point which is marked with a solid black dot. The colour of the ground truth trajectories gradually changes from black at the start to grey at the end. Results show that our method can robustly handle large occlusion in the camera view and is able to recover correct camera trajectories after drift caused by large occlusion.}
    \label{fig:cam_traj}
\end{figure*}

\begin{table*}[tb]
\resizebox{\linewidth}{!}{%
  \centering
    \begin{tabular}{|c|c|cc|cc|cc|cc|cc|cc|cc|}
    \hline
    \multicolumn{2}{|c|}{\multirow{3}[6]{*}{}} & \multicolumn{4}{c|}{Visual SLAM} & \multicolumn{10}{c|}{Visual-inertial SLAM} \\
    \cline{3-16}
    \multicolumn{2}{|c|}{}
    & \multicolumn{2}{|c}{SF$^*$ \cite{scona2018staticfusion}}
    & \multicolumn{2}{|c}{PF$^*$ \cite{long2022pnpfusion}}
    & \multicolumn{2}{|c}{VINS-Mono \cite{qin2018vins}} 
    & \multicolumn{2}{|c}{ORB-SLAM3 \cite{campos2021orb}} 
    & \multicolumn{2}{|c}{Dynamic-VINS \cite{Liu2022dynamicvins}} & 
    \multicolumn{2}{|c}{Ours$^*$} & 
    \multicolumn{2}{|c|}{Ours} \\
    \cline{3-16}
    \multicolumn{2}{|c|}{} & ATE   & RPE   & ATE   & RPE   & ATE   & RPE   & ATE & \multicolumn{1}{c|}{RPE} & ATE   & RPE   & ATE & \multicolumn{1}{c|}{RPE} & ATE & \multicolumn{1}{c|}{RPE} \\
    \hline
    1     & Static & 2.05  & 0.169 & 1.97  & 0.161  & 0.081 & 0.025 & \textbf{0.058} & \textbf{0.021} & 0.072 & 0.024 & 0.156 & 0.020 & 0.069 & 0.022 \\
    \hline
    2     & \multirow{2}[0]{*}{STLO} & 1.41  & 0.433 & 1.18  & 0.202 & $>$6.0     & 2.112     & 0.873 & 0.321 & 0.747 & 0.306 & 0.170  & 0.0483& \textbf{0.159} & \textbf{0.0471}  \\
    \cline{1-1}\cline{3-16}
    3     &       & $>$6.0     & 3.15  & 3.40     & 0.229  & 5.044     & 0.473     & 1.019 & 0.503 & 0.949 & 0.372 & 0.203 & \textbf{0.0470} & \textbf{0.081} & 0.0573 \\
    \hline
    4     & \multirow{2}[0]{*}{MTLO} & 1.76  & 0.683 & 1.78  & 0.238   & $>$6.0     & $>$6.0     & 1.934 & 0.402 & 1.973 & 0.375 & \textbf{0.151} & \textbf{0.0319} & 0.168 & 0.0342 \\
    \cline{1-1}\cline{3-16}
    5     &      & 1.66  & 0.533   & 1.54     & 0.176   & 1.234 & 0.345 & 1.584 & 0.441 & 1.181 & 0.232 & \textbf{0.178} & \textbf{0.0251} & 0.182 & 0.0289 \\
    \hline
    6     & \multirow{4}[1]{*}{LTLO} & 5.07     & 1.35   & 4.02     & 0.165  & $>$6.0     & $>$6.0    &  2.381  &  0.559  & 2.923     & 0.527     & 0.212 & 0.0771 & \textbf{0.153} & \textbf{0.0557} \\
    \cline{1-1}\cline{3-16}
    7     &       & $>$6.0     & 3.46   & 5.36     & 0.148  & $>$6.0     & 0.749    &  2.049  &  0.643  & 1.278 & 0.608 & 0.204 & 0.0844 & \textbf{0.123} & \textbf{0.0727} \\
    \cline{1-1}\cline{3-16}
    8     &       & $>$6.0     & 2.07  & 5.21     & 0.137  & $>$6.0     & 1.95     &  2.661  &  0.547  & 2.303     & 0.518     & 0.295 & \textbf{0.0420} & \textbf{0.191} & 0.0440 \\
    \cline{1-1}\cline{3-16}
    9     &       & $>$6.0     & 4.49  & 4.23     & 0.153  & $>$6.0     & 3.09     & 1.995 & 0.618 & 2.986     & 0.628     & 0.533 & \textbf{0.0665} & \textbf{0.171} & 0.0815 \\
    \hline
    \end{tabular}%
  }
  \caption{ATE (m) and RPE RMSE (m/s) for all nine collected sequences. The asterisk ($*$) symbol means either the method is unable to close loops or the loop thread is disabled. Our method outperforms all other state-of-the-art methods when the large occlusion lasts for a long period in the camera view. While in static environments, our method has comparable results to other visual-inertial SLAM methods.}
  \label{tab:camera_trajectories}%
\end{table*}%

To quantitatively evaluate camera trajectories, we estimate the ATE root-mean-square-error (RMSE) and RPE RMSE against ground truth camera trajectories (\Cref{tab:camera_trajectories}). The results are compared with SF \cite{scona2018staticfusion}, PF \cite{long2022pnpfusion}, VINS-Mono \cite{qin2018vins}, ORB-SLAM3 \cite{campos2021orb} and Dynamic-VINS \cite{Liu2022dynamicvins}. In static environments, our method achieves comparable results with other state-of-the-art visual-inertial SLAM methods. 
Both visual SLAM methods, StaticFusion and PlanarFusion, have a relative high error in static environments, because they are unable to reduce long-term drift with loop closure. In the scenario of LTLO (seq. 6-9), evaluation demonstrates that our method is able to provide accurate camera trajectories and outperforms all other methods (\Cref{fig:cam_traj}). While neither ORB-SLAM3 nor Dynamic-VINS is able track camera trajectory correctly. This is because ORB-SLAM3 is unable to remove sparse features from dynamic objects and Dynamic-VINS can only remove features from specific categories of dynamic objects like humans. In the STLO and MTLO sequences (seq. 2-5), the state-of-the-art methods have better performance compared to their performance on LTLO sequences, however, our method achieves more accurate results.

We further disable the loop closure thread of our method to quantify the accuracy improvement from the dynamic object removal. Results show that the localisation precision decreases as the duration of large occlusion increases.

\subsection{Dynamic Object Segmentation}
\begin{figure*}[htb]
\setlength{\belowcaptionskip}{-0.4cm}
\centering
\setlength{\tabcolsep}{0pt}
\newcommand{\h}{1.85cm}
\begin{tabular}{r|cccccc|}
\textbf{} & \multicolumn{6}{c|}{\textbf{Dense segmentation per frame ID}} \\
 & 500 & 790 & 994 & 1217 & 1376 & 1746 \\

RGB input\hspace{0.2cm} &  \includegraphics[height=\h,valign=m]{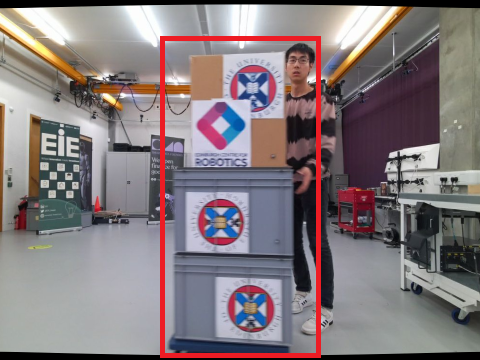}   \     & 
\includegraphics[height=\h,valign=m]{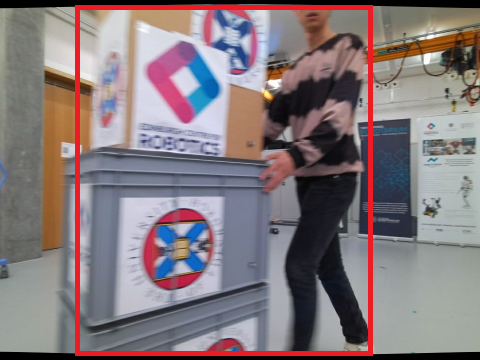}   \      & 
\includegraphics[height=\h,valign=m]{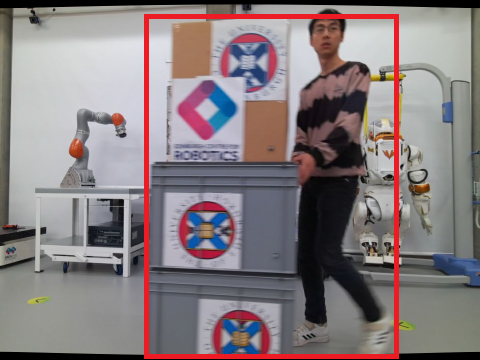}   \     &
\includegraphics[height=\h,valign=m]{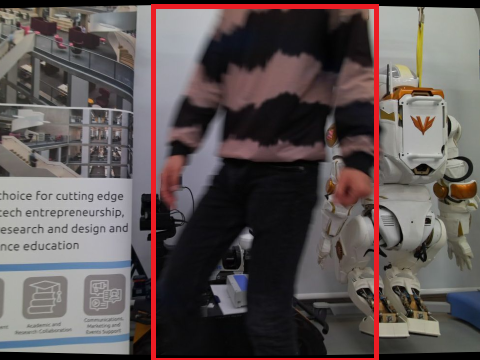}  \    & 
\includegraphics[height=\h,valign=m]{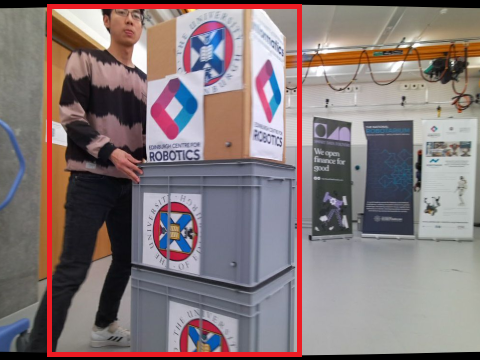}   \     & 
\includegraphics[height=\h,valign=m]{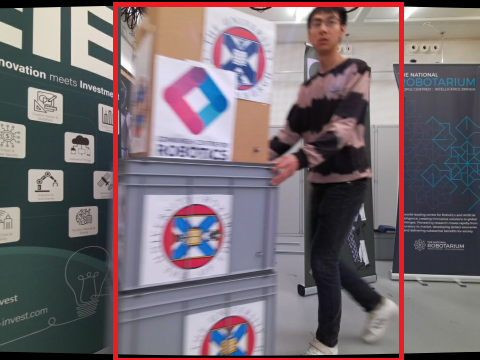} \\

SF \cite{scona2018staticfusion} \hspace{0.2cm} &  \includegraphics[height=\h,valign=m]{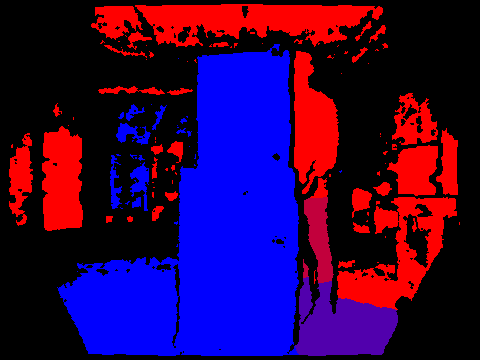}   \     & \includegraphics[height=\h,valign=m]{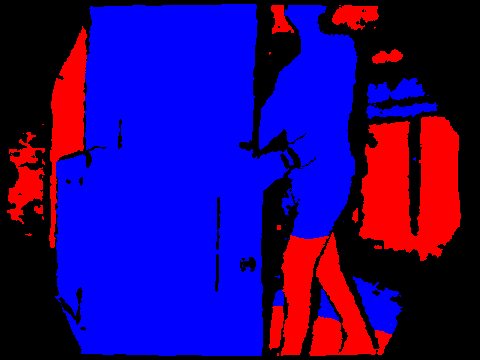}   \      & \includegraphics[height=\h,valign=m]{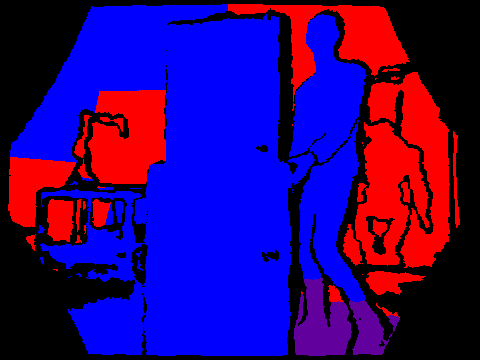}    \    &
\includegraphics[height=\h,valign=m]{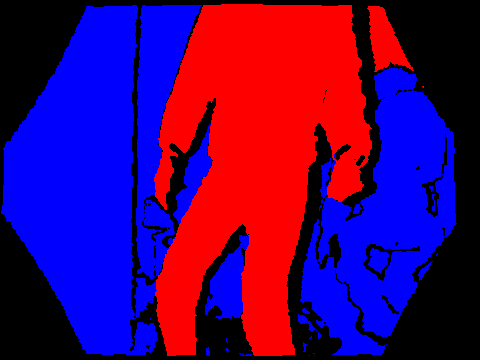}   \     & \includegraphics[height=\h,valign=m]{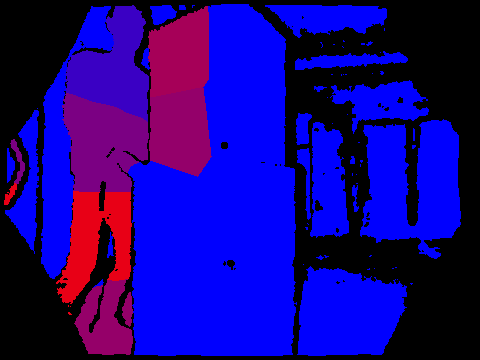}    \    & \includegraphics[height=\h,valign=m]{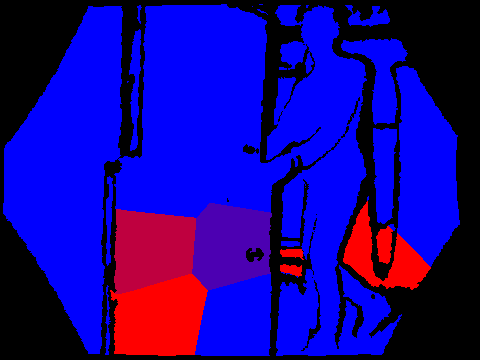}  \\

CF \cite{runz2017co} \hspace{0.2cm} & \includegraphics[height=\h,valign=m]{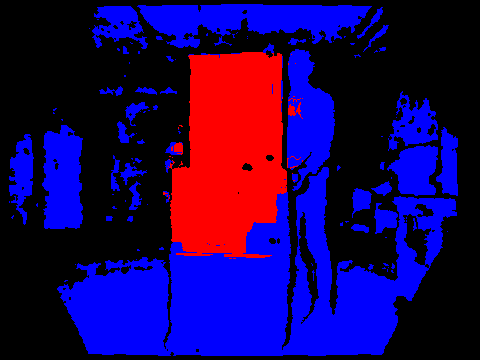}   \     & \includegraphics[height=\h,valign=m]{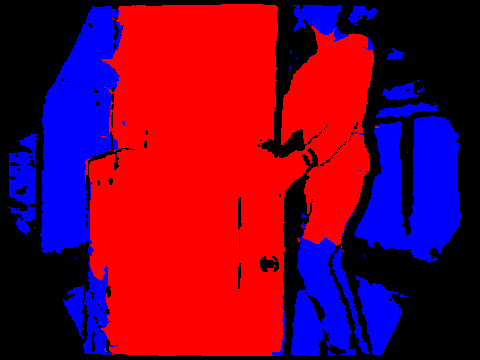}   \      & \includegraphics[height=\h,valign=m]{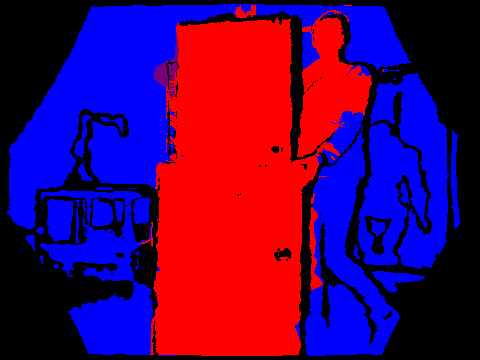}    \    &
\includegraphics[height=\h,valign=m]{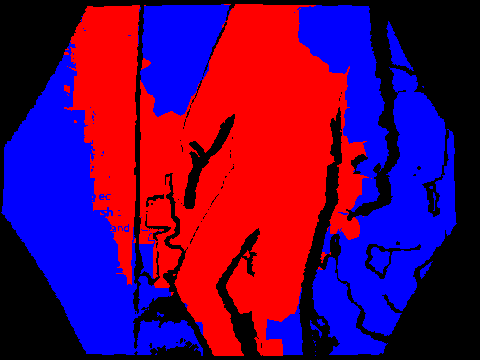}   \     & \includegraphics[height=\h,valign=m]{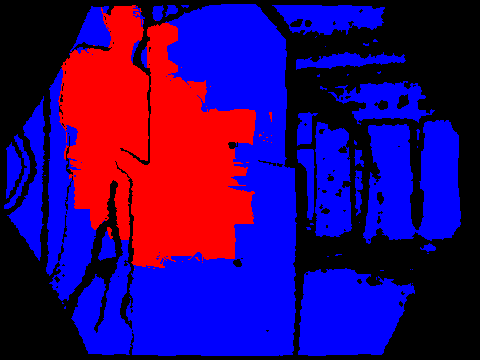}    \    & \includegraphics[height=\h,valign=m]{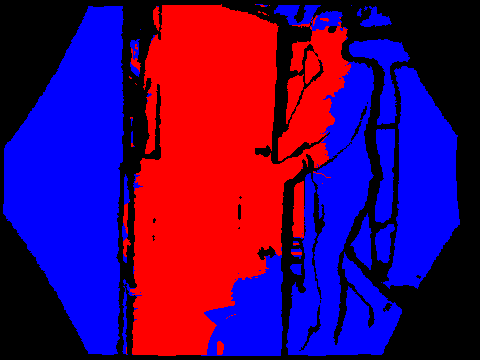}  \\

PF \cite{long2022pnpfusion} \hspace{0.2cm} &
\includegraphics[height=\h,valign=m]{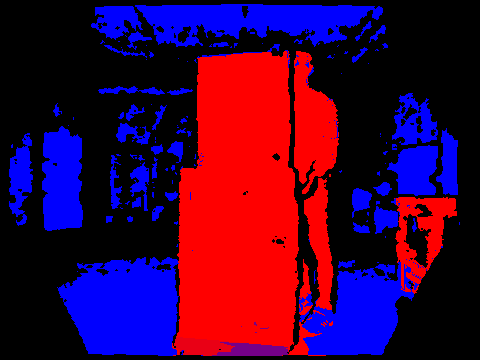}   \     & \includegraphics[height=\h,valign=m]{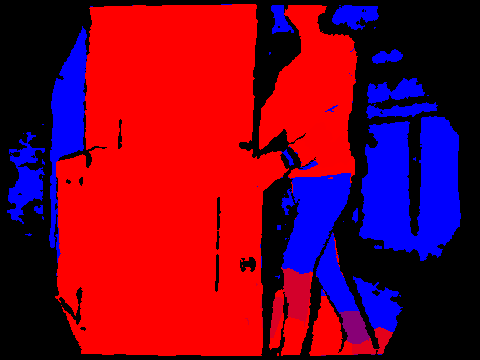}   \      & \includegraphics[height=\h,valign=m]{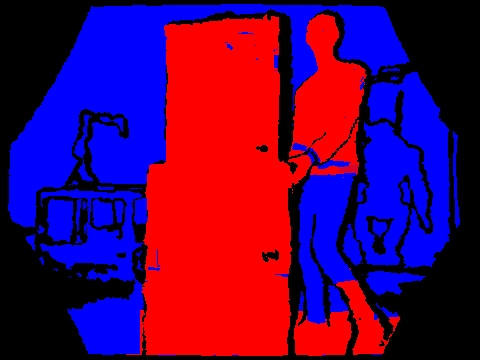}    \    &
\includegraphics[height=\h,valign=m]{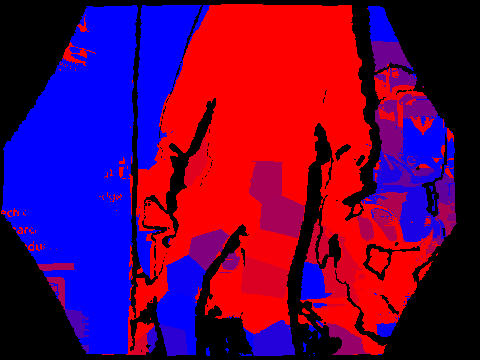}   \     & \includegraphics[height=\h,valign=m]{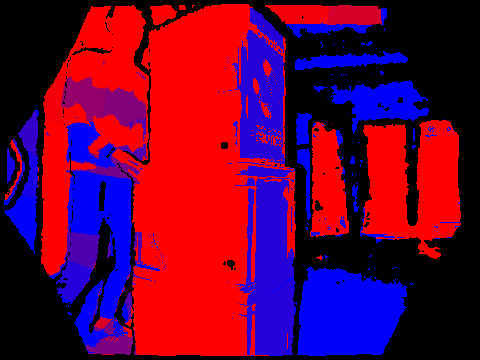}    \    & \includegraphics[height=\h,valign=m]{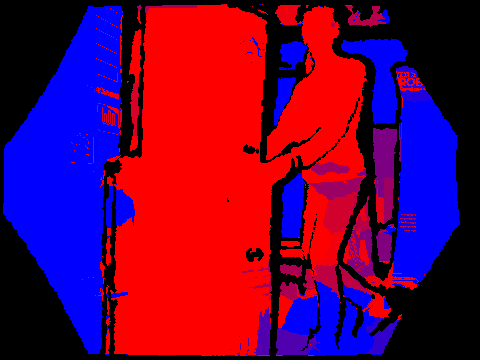}  \\

Ours \hspace{0.2cm}    & \includegraphics[height=\h,valign=m]{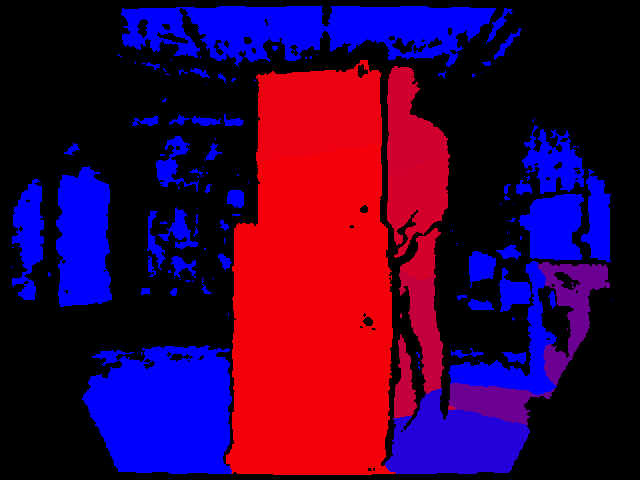}   \     & \includegraphics[height=\h,valign=m]{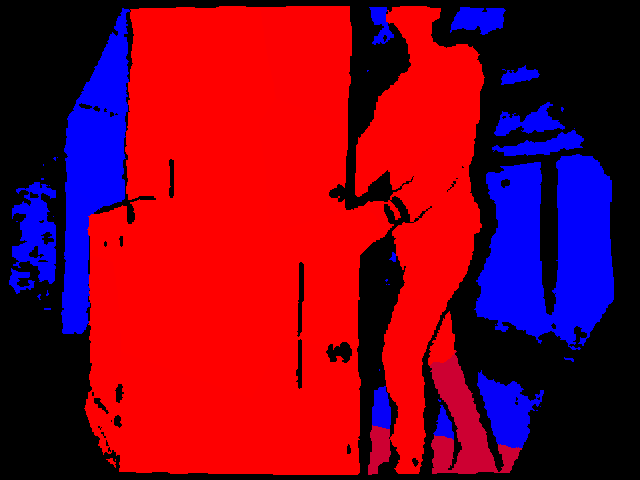}   \      & \includegraphics[height=\h,valign=m]{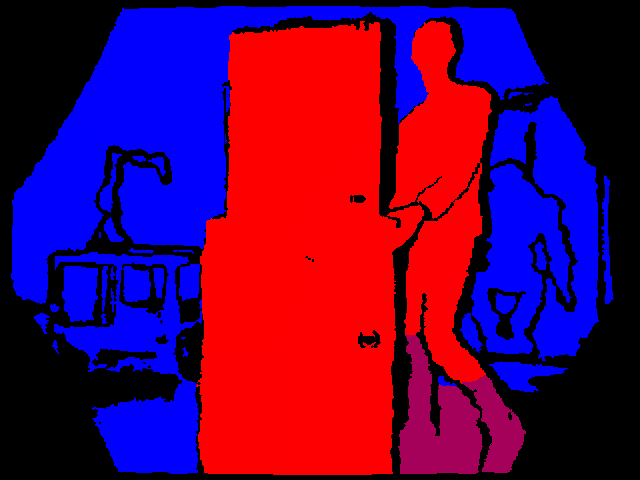}    \    &
\includegraphics[height=\h,valign=m]{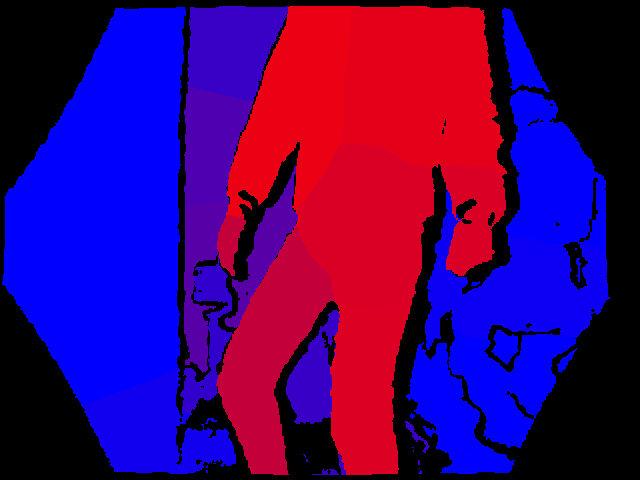}   \     & \includegraphics[height=\h,valign=m]{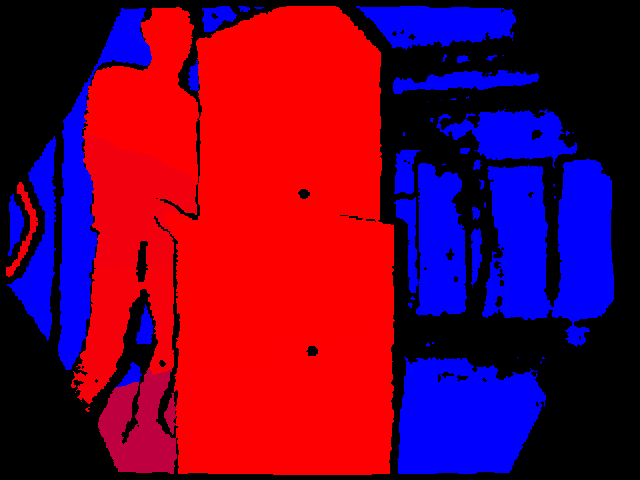}    \    & \includegraphics[height=\h,valign=m]{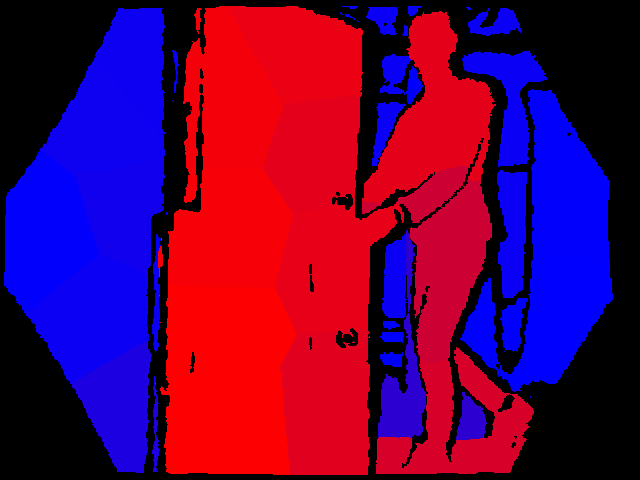}  \\
\end{tabular}
\caption{Segmentation result of the static background (blue) and dynamic objects (red) in \textit{seq7}. In the first row, we show the input RGB images and their corresponding time frame ID. The dynamic objects are manually highlighted by red rectangles for better visualisation. Results show that only our method can provide a consistent segmentation of objects that cause large occlusion for a long period of time. In contrast, both SF and CF are unable to segment the dynamic objects correctly, while the segmentation performance of PF is not persistent over time.
}
\label{fig:segmentation}
\end{figure*}

\Cref{fig:segmentation} shows the static/dynamic segmentation results of \textit{seq7} and the majority of camera view is continuously occluded for more than 40 seconds from the time frame 500 to 1746. We compare our segmentation results against StaticFusion (SF) \cite{scona2018staticfusion}, Co-Fusion (CF) \cite{runz2017co} and PlanarFusion (PF) \cite{long2022pnpfusion}. Results show that SF is unable to detect dynamic objects when they occupy the major proportion of the visual input. In addition, the segmentation provided by CF is incomplete and parts of dynamic objects are classified as the static background. Although PF achieves better segmentation results than SF and CF, it is unable to provide accurate results consistently over a long period of time because only two consecutive image frames are used. Our method takes advantage of local optimisation over multiple keyframes and can therefore detect multiple dynamic objects that cause long-term large occlusion.

\subsection{Background Reconstruction}
\begin{figure*}[htb]
\setlength{\belowcaptionskip}{0cm}
\centering
\setlength{\tabcolsep}{0pt}
\newcommand{\h}{0.24\linewidth}
\begin{tabular}{c|c|c|c}

\includegraphics[width=\h]{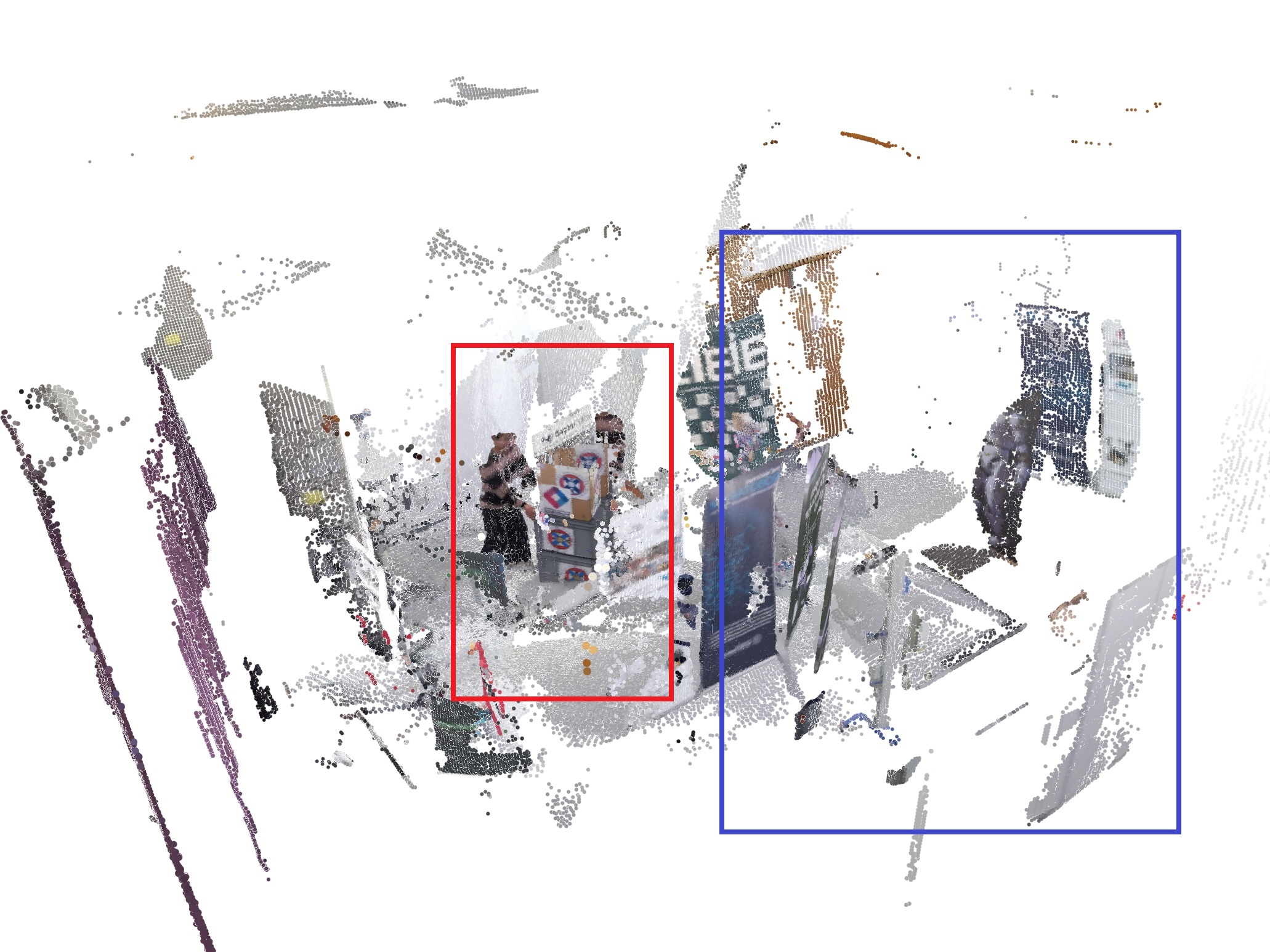}       & 
\includegraphics[width=\h]{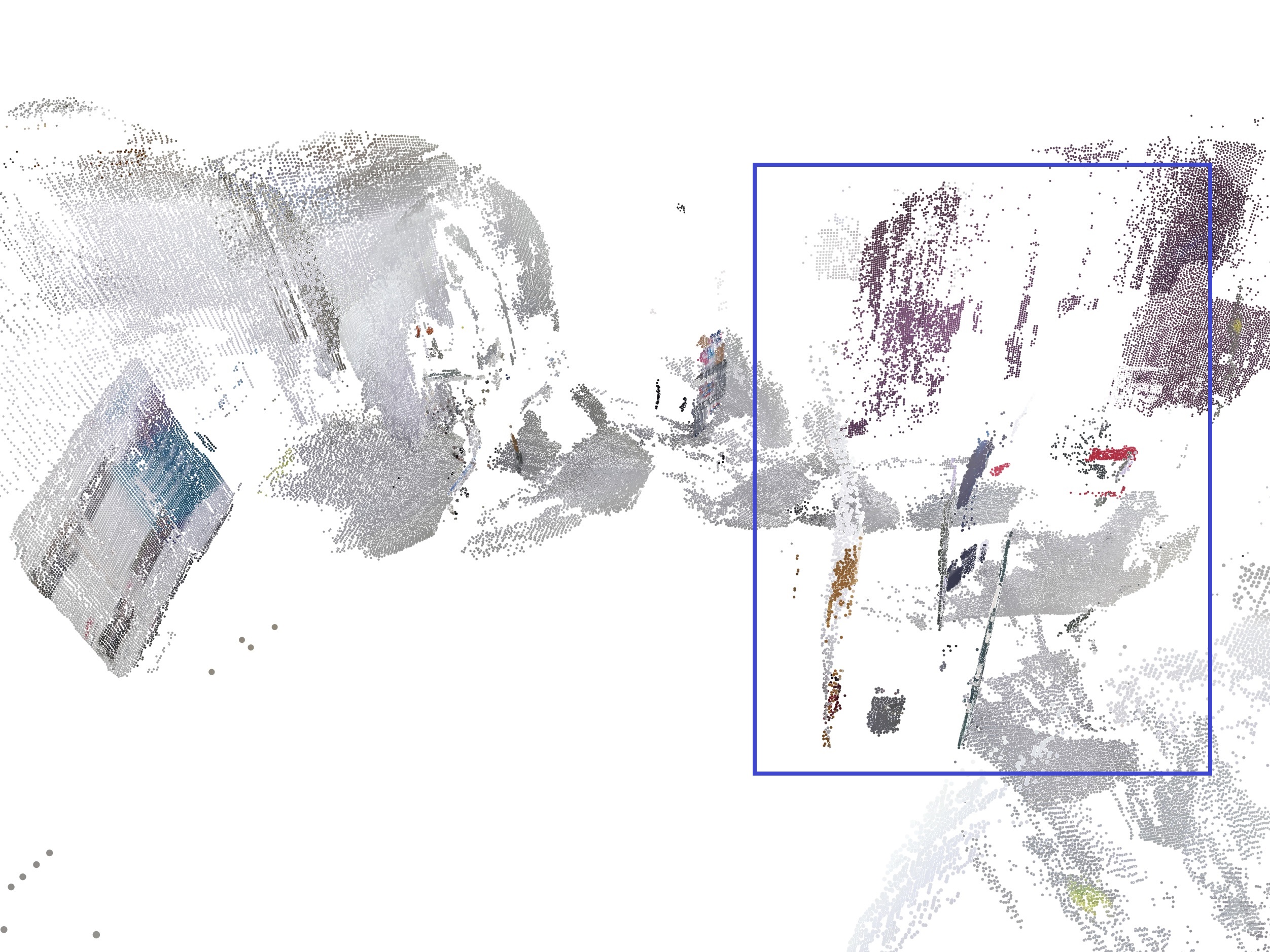}       &
\includegraphics[width=\h]{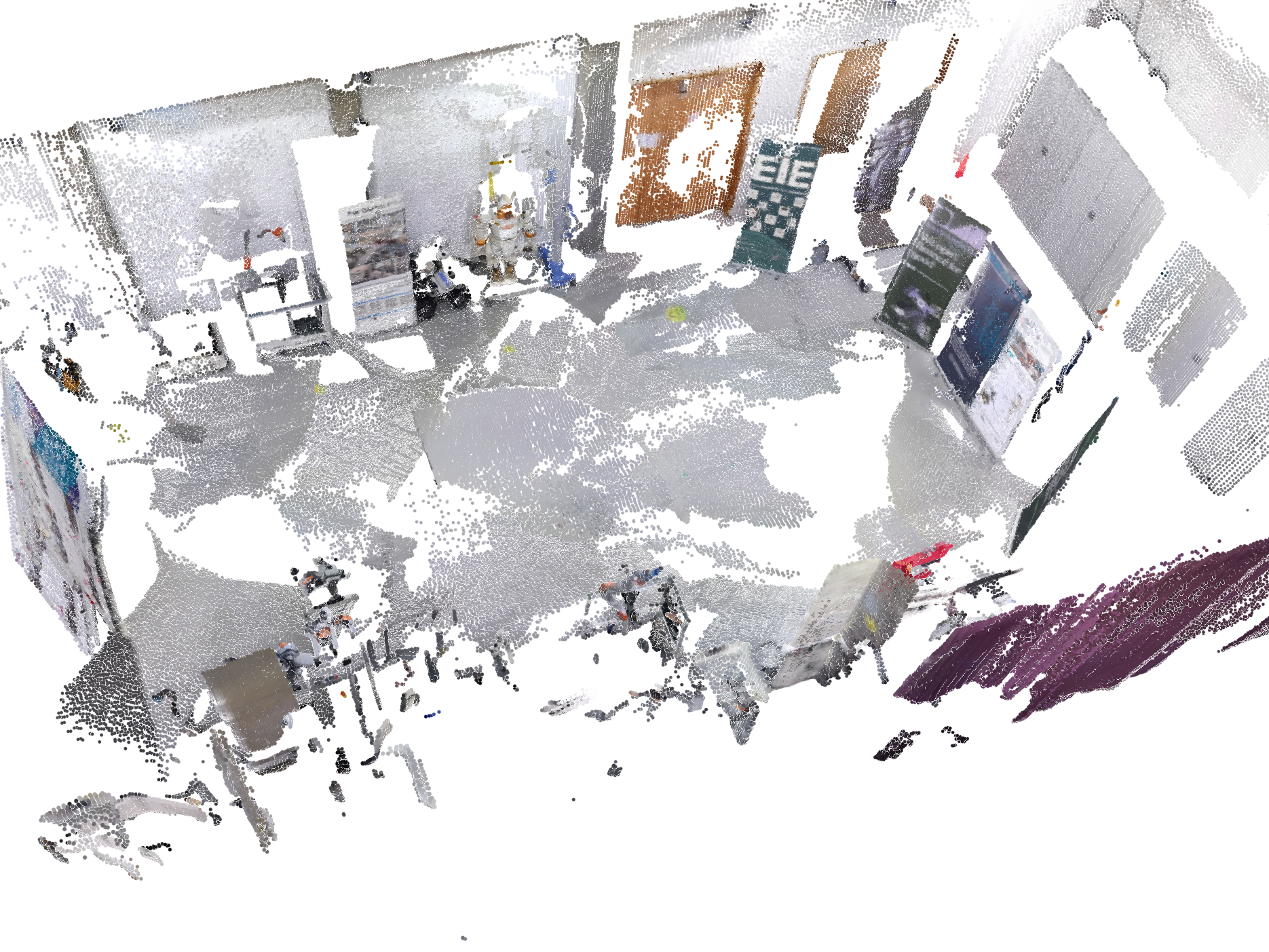}    & 
\includegraphics[width=\h]{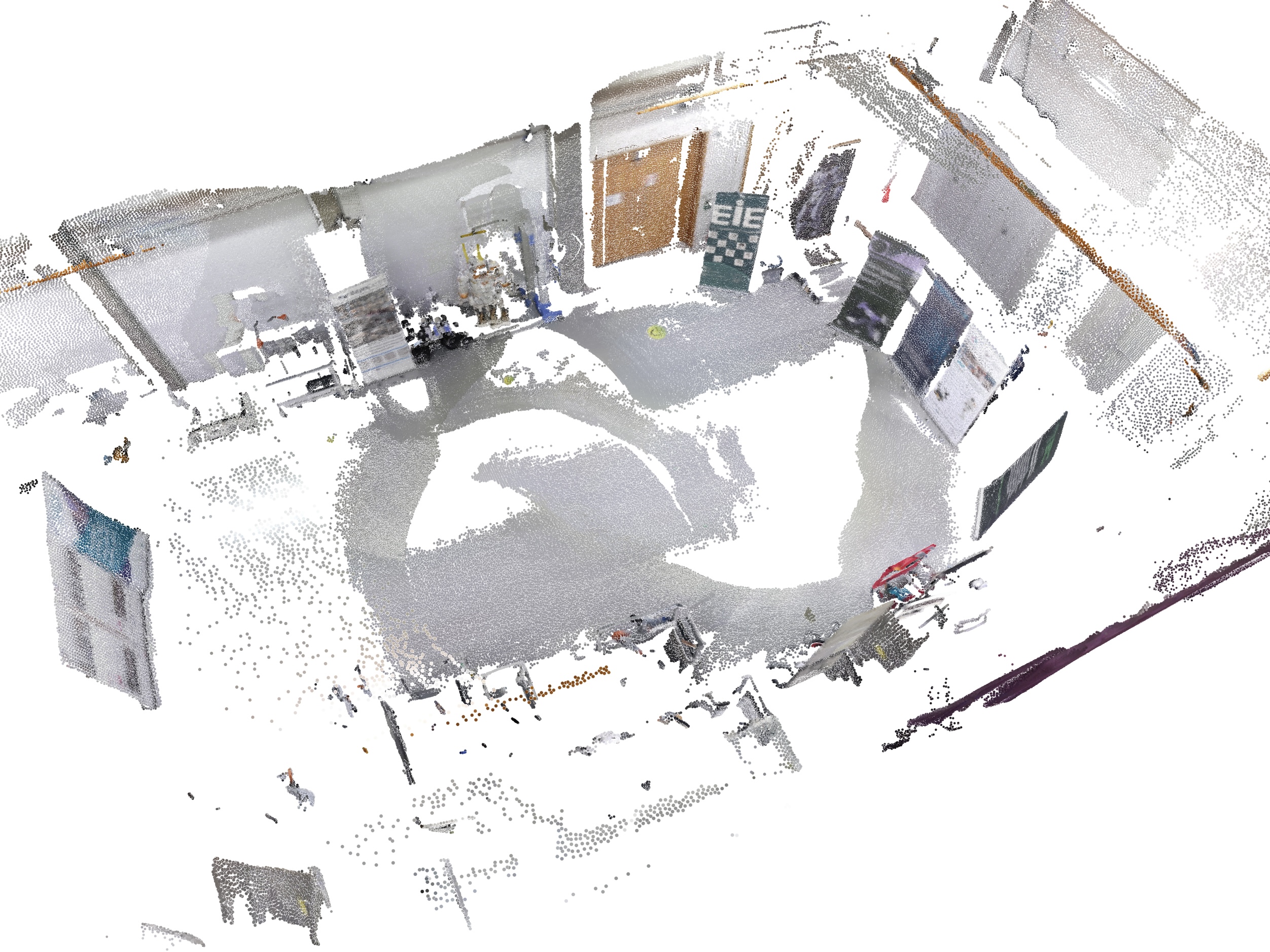} \\
SF & PF & Ours & Ground truth

\end{tabular}
\caption{Reconstruction results of the RGB-D sequence 7. We highlight the dynamic objects with red rectangle and the dislocation of static objects with blue rectangle. SF can neither remove dynamic objects nor estimate camera trajectory correctly. In contrast, both PF and our method can detect dynamic object but PF is unable to accurately localise camera after removal of dynamic objects. }
\label{fig:reconstruction}
\end{figure*}

We qualitatively compare our dense background reconstruction results with SF and PF (\Cref{fig:reconstruction}). We reconstruct the background after processing the whole sequence with the estimated dense segmentation and camera trajectory.  The ground truth model is acquired by mapping the static environment to the ground truth camera trajectory. Results show that SF maps dynamic objects into the static background model, while PF is able to remove dynamic objects from the model. However, the dislocation of static objects indicates that neither SF nor PF is able to estimate camera ego-motion correctly. Based on consistent dynamic object segmentation and camera localisation, our method outperforms other methods and provides a correct background reconstruction.

\section{Conclusion}
This work presents a novel RGB-D-inertial SLAM method that is robust to multiple undefined dynamic objects that cause long-term large occlusion. Our proposed robust visual-inertial bundle adjustment can simultaneously estimate the dense segmentation of dynamic objects and localise camera with a combination of dense and sparse features. The dense segmentation can be used to reconstruct the background. The detailed evaluation demonstrates that our proposed approach outperforms other state-of-the-art methods in terms of the dense object segmentation, camera localisation and background reconstruction in the presence of long-term large occlusion. 

However, our current method can fail to detect dynamic objects in outdoor or texture-less environments and is unable to track or model dynamic objects. Our future work will aim to address long-term object tracking and modelling, building on the gains realised by the proposed method. We also plan to extend the current method to large-scale outdoor environments.

\section*{ACKNOWLEDGEMENT}

This research is supported by the EU H2020 project Enhancing Healthcare with Assistive Robotic Mobile Manipulation (HARMONY, 9911237), the Shenzhen Institute of Artificial Intelligence and Robotics for Society (AIRS), the Kawada Robotics Corporation and the Alan Turing Institute. 


\bibliographystyle{IEEEtran.bst}
\bibliography{IEEEabrv,mybibfile}

\end{document}